# Compactified Horizontal Visibility Graph for the Language Network

D.V. Lande[1,2], A.A.Snarskii[1,2]

[1]Institute for information recording, NAS Ukraine, Kiev, Ukraine
[2]National Technical University "Kiev Polytechnic Institute", Kiev, Ukraine

*A compactified horizontal visibility graph for the language network is proposed. It was found that the networks constructed in such way are scale free, and have a property that among the nodes with largest degrees there are words that determine not only a text structure communication, but also its informational structure.*

**Key words:** language network, complex network, scale-free network, visibility graph.

Construction of networks with text elements, words, phrases or fragments of natural language as nodes in some cases allows to detect the structural elements of the text critical for its connected structure and find informationally significant elements, as well as words that are secondary for understanding of the text. Such networks may also be used to identify unconventional text components, such as collocations, supra-phrasal units [1], as well as for finding similar fragments in different texts [2].

There is a multitude of approaches to constructing networks from the texts (so-called language networks) and different ways of interpreting nodes and links, which causes, accordingly, different representation of such networks. Nodes are connected if corresponding words are either adjacent in the text [3, 4], or are in a single sentence [5], or are syntactically [6, 7] or semantically [8, 9] connected.

At the intersection of digital signal processing (DSP) theory and complex network theory there are several ways of constructing networks from the time series, among those are visibility graph construction methods (see survey [10]), namely the horizontal visibility graph (HVG) [11,12]. Based on these approaches, networks can also be constructed from texts in which numeric values are assigned in some manner to each word or phrase. The examples of functions assigning a number to a word are: ordinal number of a unique word in a text, length of the word, "weight" of the word in a text, e.g., generally accepted TFIDF metric (canonically, a product of the term frequency in a text fragment and a binary logarithm of the inverse number of text fragments containing this word – inverse document frequency) or its modifications [13, 14] and other word weight estimates.

In this paper, the standard deviation estimate of word weight is used for constructing word networks [15]. If all the words in the text of $N$ words are numbered in succession (let $n=1,...N$ be the ordinal number of the word in a text, the word position), layout of a certain word $A$ can be designated as $A_k(n)$, where $k=1,2,...,K$ denotes the number of occurrence of this word in a text, and $n$ is a position of this word in a text. For example, $A_3(50)$ means that the third occurrence of the word $A$ has position 50 in the text.

The distance between successive occurrences of the word in these terms would be $\Delta A_k = A_{k+1}(m) - A_k(n) = m-n$, where $m$ and $n$ are the positions of the $k+1$-th $k$-th occurrences of the word $A$ in the text, respectively.

Standard deviation estimate proposed in [15] is calculated as follows:

$$\sigma_A = \frac{\sqrt{\langle \Delta A^2 \rangle - \langle \Delta A \rangle^2}}{\langle \Delta A \rangle}, \tag{1}$$

where $\langle \Delta A \rangle$ is a mean value of the sequence $\Delta A_1, \Delta A_2, ..., \Delta A_K$, $\langle \Delta A^2 \rangle$ is a mean value of $\Delta A_1^2, \Delta A_2^2, ..., \Delta A_K^2$, and $K$ is a number of occurrences of the word $A$ in the text.

As opposed to other series examined in DSP theory, the series of numerical values assigned to words are transformed into horizontal visibility graphs (HVG), where each node not only has a corresponding numerical value, but also the corresponding word itself.

The process of constructing the language network using HVG consists of two stages. At the first stage, the traditional HVG is constructed [16]. To do that a series of nodes is put on the horizontal axis, where each node corresponds to a word in order of occurrence in the text, and standard deviation estimates are put on the vertical axis (visually a histogram, see fig. 1). There is a connection between nodes if they are in “line of sight” with each other, i.e., if they can be connected by a horizontal line that does not cross any other histogram bar. This (geometric) criterion can be written down as follows, according to [10,11]: the two nodes (words), e.g., $B_3(n)$ and $C_7(m=n+5)$, are connected if (see fig. 1)

$$\sigma_n, \sigma_m > \sigma_p \text{, for all } n < p < m. \tag{2}$$

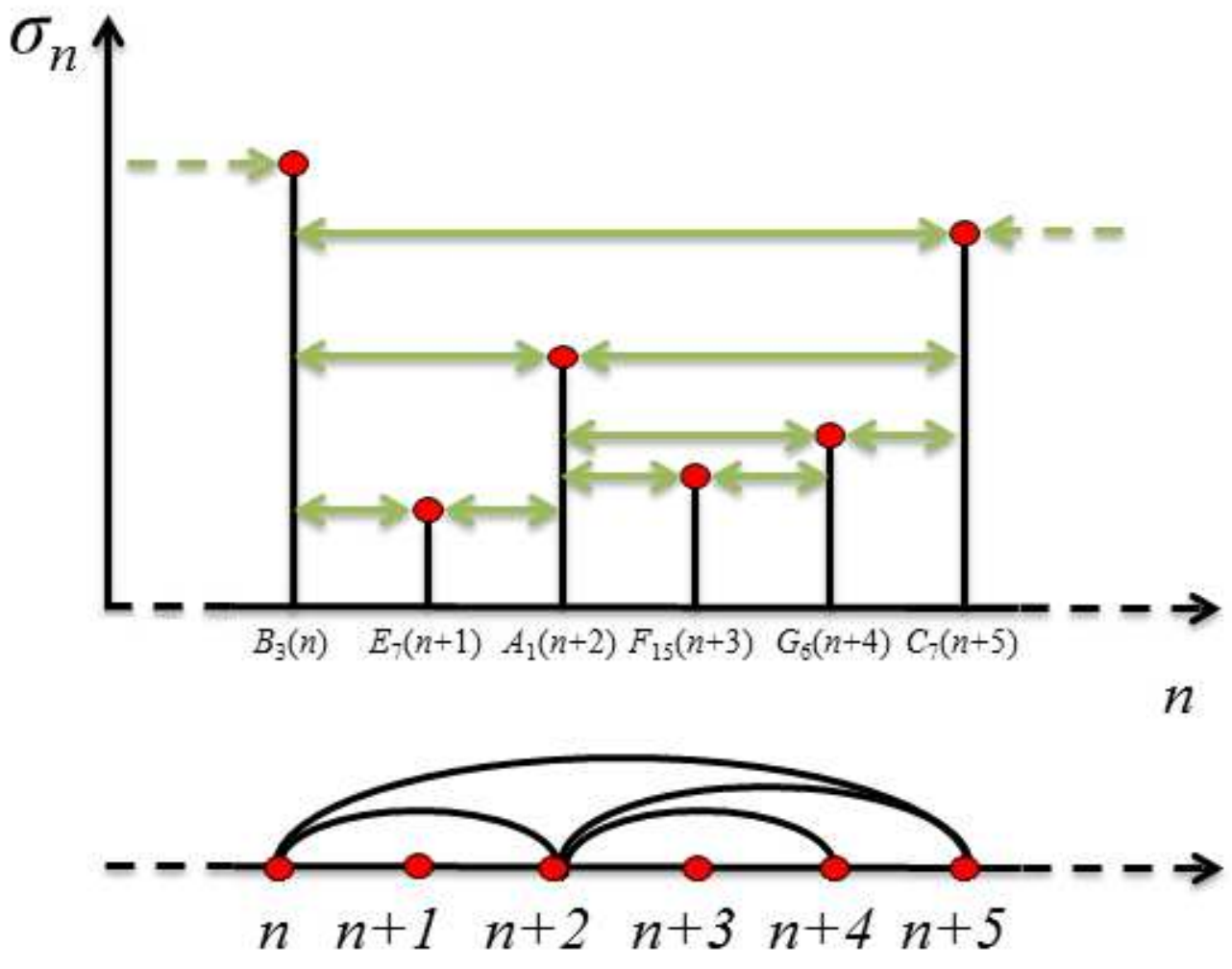

Figure 1. An example of HVG construction

The process of constructing can be algorithmized. For example, in figure 1 the word node $A_1(n+2)$ is considered incident (and is connected with edges) to the words $B_3(n)$ and $C_1(n+5)$, $B_3(n)$ being the closest word to the left of $A_1(n+2)$ with a standard deviation estimate $\sigma_n = \sigma_B$ greater than that of the word *A*: $\sigma_{n+2} = \sigma_A$, and $C_7(m=n+5)$ being the closest word to the right of $A_1(n+2)$, for which $\sigma_m > \sigma_A$.

At the second stage, the derived network is compactified. All the nodes corresponding to a single word, e.g., the word $A$, are combined into a single node (naturally, occurrence numbers and positions of the words are lost). The connections of theses nodes are also combined. Note that there is no more than one edge left between any pair of nodes, multiple connections are removed (see fig. 2).

This means, in particular, that the degree (number of connections) of the node $A$ does not exceed the sum of degrees $\sum_k A_k(n)$. As a result, the new network of words – *compactified horizontal visibility graph* (CHVG) – is constructed (fig. 2).

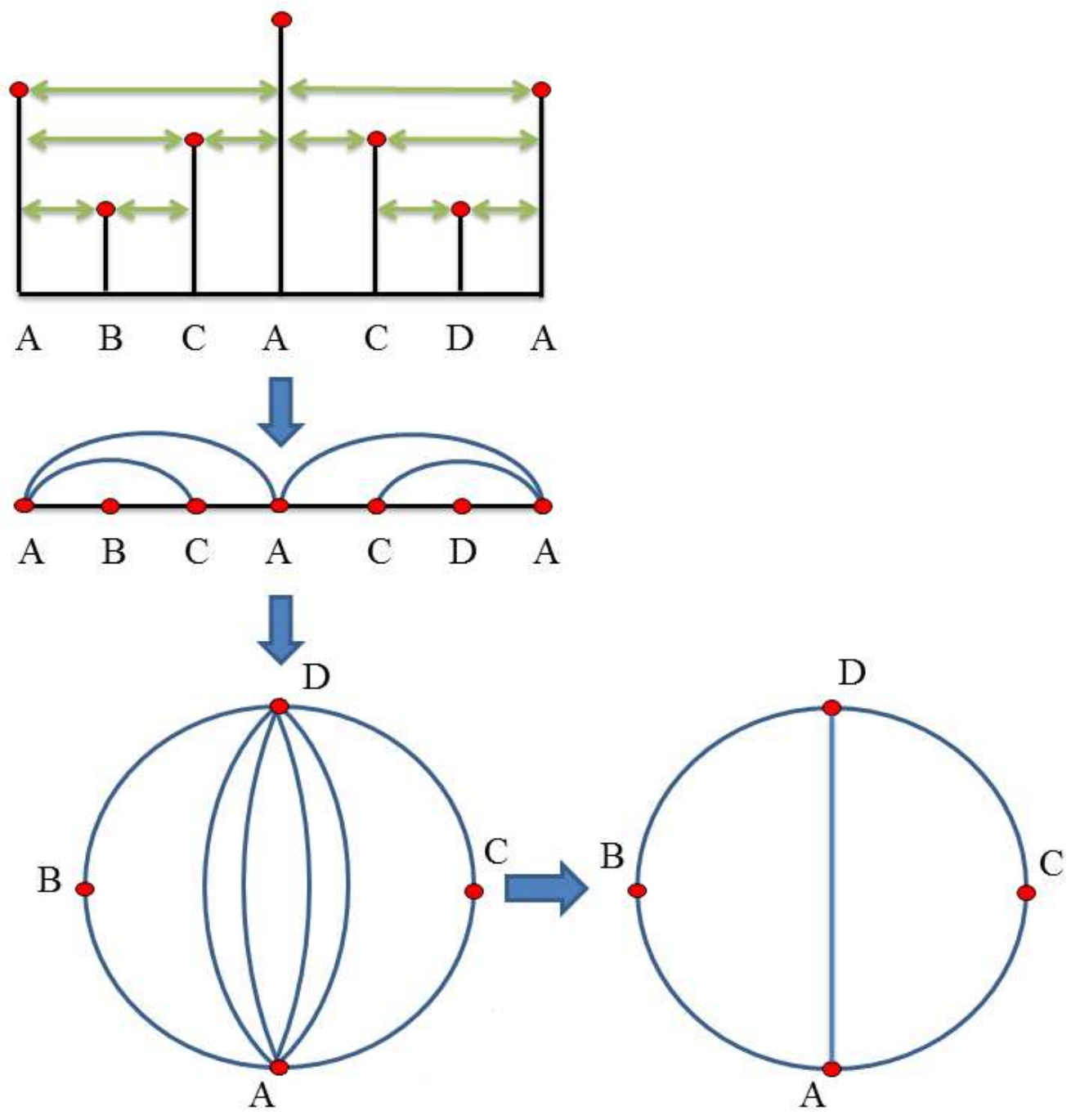

Figure 2. Two stages in construction of CHVG

Texts used for CHVG construction were the novels “The Master and Margarita” (original version) by Mikhail Bulgakov and “Moby-Dick; or, The Whale” by Herman Melville, as well as arrays of news information from the Web.

For all CHVG networks of words described here, the degree distribution is close to power law (fig. 3), i.e., these networks are scale free.

For comparison, was studied for the simplest language networks, where during the first stage of the network construction adjacent words were connected, and, at the second stage, the network was compactified. It is obvious that the weight of a node in such network corresponds to the word frequency, and the distribution of these weights follows the Zipf law [18]. The most connected are the nodes corresponding to the most frequently occurring words – conjunctions, prepositions, etc., which are very important for the text coherence, but are of little interest for the aspect of informational structure.

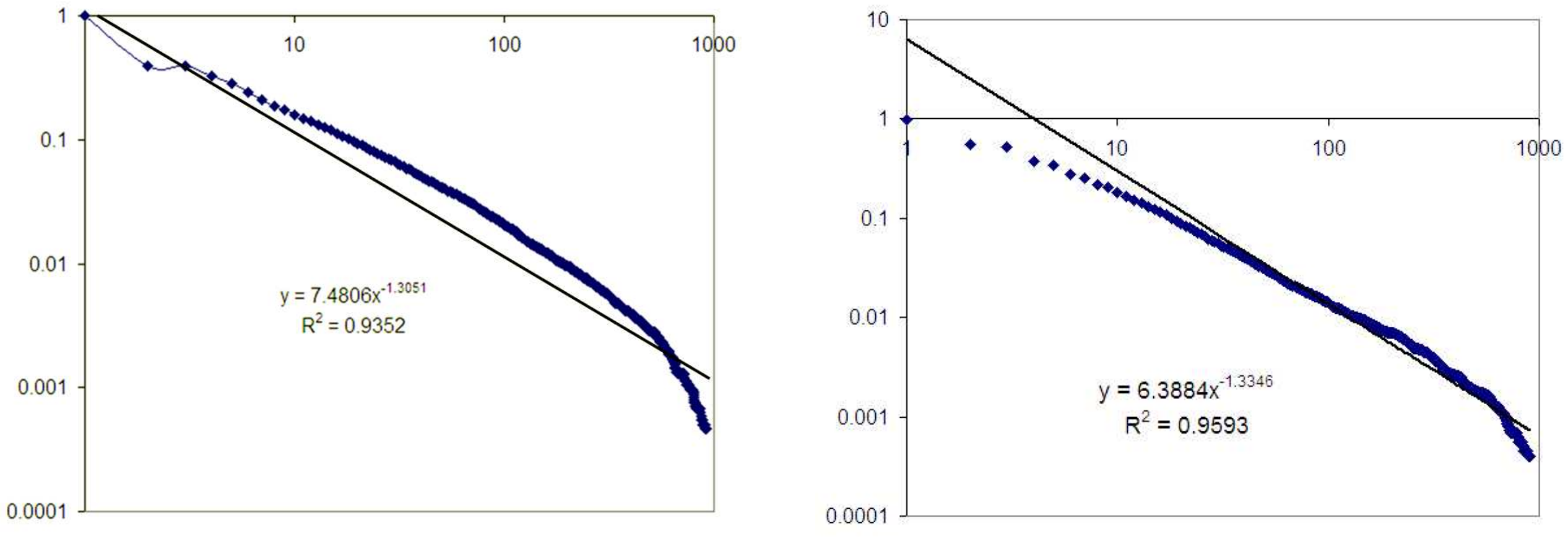

a) b)

Figure 3. Node degree distribution (log-log scale) of CHVG constructed from “The Master and Margarita” (a) and “Moby-Dick; or, The Whale” (b). Horizontal axis contains node degrees $k$, vertical axis shows the values $1 - F(k)$, where $F(k)$ is a distribution function of node degrees

Among the nodes with largest degrees, alongside with personal pronouns and other function words (particles, prepositions, conjunctions, etc.), are the words, which determine the informational structure of the text [16, 17].

Let $\Psi$ be a set of $N$ different words (in our case $N$ = 100) corresponding to the largest-weight nodes of the aforementioned simple language network, and let $\Lambda$ be a set of words corresponding to the largest-weight nodes of the CHVG. Then the set $\Omega = \Lambda \setminus \Psi$ will contain informational words, which are also important for the text coherence. Appendix gives juxtaposition of the top 100 largest-weight nodes for the two types of language networks constructed from the novels “The Master and Margarita” by Michael Bulgakov and “Moby-Dick; or, The Whale” by Herman Melville.

In particular, the $\Omega$ set of the CHVG built from “Мастер и Маргарита” contains such words as Иван, Мастер, Варенуха, Берлиоз, Бегемот, Римский, профессор, Левий, Иешуа.

The following results were obtained from studying the language networks:

1. An algorithm compactified horizontal visibility graph (CHVG) was proposed.
2. Language networks were built from different texts based on series of standard deviation estimates and CHVG.

3. In CHVG obtained from literary works, among the largest-degree nodes there are words responsible not only for the coherence of the text, but also for its informational structure. They reflect the meaning of the mentioned texts.

## Appendix

Table 1. Juxtaposition of the top 100 largest-weight nodes of the word networks constructed from Bulgakov's "The Master and Margarita"*

| Simple network | | CHVG | |
|---|---|---|---|
| **Weight** | **Word** | **Weight** | **Word** |
| 5724 | И | 14724 | И |
| 3591 | В | 12880 | В |
| 2235 | НА | 8069 | НЕ |
| 1893 | НЕ | 7550 | НА |
| 1616 | С | 6511 | ЧТО |
| 1396 | ЧТО | 6050 | ОН |
| 1204 | ОН | 5225 | ТО |
| 1081 | А | 5224 | Я |
| 979 | ЕГО | 5105 | С |
| 936 | ТО | 4518 | *МАРГАРИТА* |
| 936 | КАК | 3642 | ЕГО |
| 899 | НО | 3396 | А |
| 809 | К | 3009 | К |
| 760 | Я | 2996 | КАК |
| 709 | ИЗ | 2848 | **ИВАН** |
| 680 | ПО | 2847 | ОНА |
| 634 | ЗА | 2562 | ИЗ |
| 555 | ОТ | 2509 | ВЫ |
| 553 | У | 2441 | *ПРОКУРАТОР* |
| 534 | ЭТО | 2317 | ЗА |
| 521 | ВСЕ | 2313 | ПО |
| 520 | ЖЕ | 2206 | БЫЛО |
| 514 | ОНА | 2076 | ЭТО |
| 484 | МАРГАРИТА | 2057 | НО |
| 460 | ЕЕ | 2000 | У |
| 409 | БЫЛО | 1989 | О |
| 403 | ПОД | 1940 | ЕЕ |
| 403 | БЫЛ | 1914 | ВСЕ |
| 400 | ТАК | 1904 | *КОРОВЬЕВ* |
| 382 | ВЫ | 1859 | *ВОЛАНД* |
| 379 | УЖЕ | 1815 | БЫ |
| 375 | ЕМУ | 1761 | БЫЛ |
| 333 | БЫ | 1721 | *КОТ* |
| 328 | О | 1696 | ТАК |
| 321 | ТУТ | 1693 | *АЗАЗЕЛЛО* |
| 313 | ТОЛЬКО | 1687 | ЖЕ |
| 307 | ЕЩЕ | 1602 | ПОД |
| 297 | ТЫ | 1568 | ТЫ |
| 297 | МНЕ | 1439 | *ПИЛАТ* |
| 281 | НИ | 1418 | ОТ |
| 281 | МЕНЯ | 1374 | **БЕРЛИОЗ** |
| 281 | ДА | 1337 | НИ |
| 277 | ЭТОГО | 1323 | МНЕ |
| 276 | ИВАН | 1321 | МЕНЯ |
| 258 | ГДЕ | 1315 | ЕМУ |
| 254 | ЧТОБЫ | 1208 | ДА |
| 254 | ОЧЕНЬ | 1179 | ТУТ |
| 250 | КОГДА | 1147 | ВОТ |
| 250 | ДО | 1095 | НЕТ |
| 241 | НЕТ | 1030 | ТОЛЬКО |

| Simple network | | CHVG | |
|---|---|---|---|
| **Weight** | **Word** | **Weight** | **Word** |
| 237 | ЭТОТ | 1020 | *ЧЕЛОВЕК* |
| 222 | КОТ | 1007 | ВАС |
| 219 | ПРОКУРАТОР | 978 | СКАЗАЛ |
| 219 | ГЛАЗА | 961 | ЭТОГО |
| 215 | СО | 944 | **ГОСТЬ** |
| 213 | ВАС | 919 | ГДЕ |
| 212 | ИЛИ | 905 | **ВАРЕНУХА** |
| 210 | ВОТ | 886 | **МАСТЕР** |
| 209 | СОВЕРШЕННО | 871 | **НИКАНОР** |
| 207 | ЧЕЛОВЕК | 866 | **БУФЕТЧИК** |
| 206 | ЛИ | 861 | УЖЕ |
| 206 | КОРОВЬЕВ | 825 | ТЕПЕРЬ |
| 204 | ТЕПЕРЬ | 815 | ЕЩЕ |
| 199 | АЗАЗЕЛЛО | 807 | ЧТОБЫ |
| 197 | ИХ | 805 | **ИВАНОВИЧ** |
| 193 | СКАЗАЛ | 803 | НУ |
| 187 | НАД | 798 | **СТЕПА** |
| 184 | ВАМ | 790 | НАД |
| 183 | СЕБЯ | 766 | ВАМ |
| 183 | ОНИ | 761 | ВО |
| 183 | КТО | 740 | **РИМСКИЙ** |
| 182 | БЫЛА | 738 | ОЧЕНЬ |
| 177 | ПЕРЕД | 724 | ОТВЕТИЛ |
| 175 | ТОТ | 722 | СО |
| 172 | ЧЕРЕЗ | 720 | КОГДА |
| 171 | БЫЛИ | 719 | НИЧЕГО |
| 166 | ВО | 671 | **МАРГАРИТЕ** |
| 165 | ВОЛАНД | 663 | *ЛИЦО* |
| 165 | НЕГО | 657 | **ПРОФЕССОР** |
| 162 | ТОГДА | 656 | ЛИ |
| 157 | ОТВЕТИЛ | 652 | **ИВАНА** |
| 157 | ЛИЦО | 651 | ЧЕРЕЗ |
| 156 | ДАЖЕ | 649 | МЫ |
| 153 | ВРЕМЯ | 644 | ВРЕМЯ |
| 150 | СЕЙЧАС | 641 | ДО |
| 149 | ЧЕМ | 636 | ОНИ |
| 149 | ПИЛАТ | 633 | НЕГО |
| 147 | ПРИ | 623 | ЭТОТ |
| 147 | ПОСЛЕ | 619 | ПОСЛЕ |
| 147 | ЕЙ | 612 | **МАРГАРИТЫ** |
| 145 | ОПЯТЬ | 609 | **БЕГЕМОТ** |
| 144 | НУ | 607 | ИХ |
| 141 | КАКОЙ | 598 | ЧЕМ |
| 139 | ЗДЕСЬ | 590 | ЕЙ |
| 139 | МЫ | 588 | ТОГО |
| 138 | НИЧЕГО | 577 | **ЛЕВИЙ** |
| 138 | КОНЕЧНО | 575 | СЕБЯ |
| 137 | ТАМ | 575 | **АФРАНИЙ** |
| 137 | БЕЗ | 569 | **ИЕШУА** |
| 136 | ТОГО | 568 | КАКОЙ |

* The words present in the first one hundred of CHVG nodes but absent from the first one hundred of simple network nodes are in bold. The most informationally significant words from the CHVG top 100, which are also present in simple network top 100, are in italics.

Table 2. Juxtaposition of the top 100 largest-weight nodes of the word networks constructed from Melville's "Moby-Dick; or, The Whale"*

| Simple network | | CHVG | |
|---|---|---|---|
| **Weight** | **Word** | **Weight** | **Word** |
| 6612 | THE | 41291 | THE |
| 5589 | AND | 23567 | OF |
| 4257 | OF | 17704 | I |
| 3083 | A | 16585 | A |
| 2862 | TO | 16577 | AND |
| 2730 | IN | 14853 | HIS |
| 2050 | THAT | 11976 | IS |
| 1915 | HIS | 11961 | TO |
| 1568 | BUT | 11582 | HE |
| 1524 | IT | 11431 | WAS |
| 1400 | HE | 10956 | IN |
| 1341 | WITH | 9883 | **WHALE** |
| 1301 | FOR | 9516 | THAT |
| 1281 | I | 9244 | IT |
| 1248 | AS | 7483 | AS |
| 1166 | IS | 7224 | YOU |
| 1152 | WAS | 6640 | **AHAB** |
| 1148 | THIS | 6457 | HIM |
| 1086 | ALL | 5727 | BE |
| 1008 | BY | 4867 | BY |
| 977 | SO | 4753 | THIS |
| 924 | OR | 4747 | ALL |
| 887 | AT | 4647 | WITH |
| 847 | FROM | 4578 | ME |
| 832 | ON | 4511 | BUT |
| 796 | NOW | 4403 | HAD |
| 784 | NOT | 4182 | YE |
| 733 | WERE | 4147 | THEIR |
| 721 | THERE | 4143 | FROM |
| 713 | ONE | 4038 | FOR |
| 703 | HIM | 3921 | MY |
| 697 | THEIR | 3645 | WERE |
| 694 | YOU | 3618 | NOT |
| 684 | BE | 3405 | AT |
| 671 | LIKE | 3352 | **BOAT** |
| 653 | THEY | 3289 | *SHIP* |
| 643 | THEN | 3276 | ON |
| 614 | ARE | 3238 | ARE |
| 609 | MY | 3113 | THEY |
| 597 | HAD | 3104 | OR |
| 596 | WHICH | 3077 | *STUBB* |
| 594 | WHALE | 3077 | **QUEEQUEG** |
| 581 | SOME | 3052 | NOW |
| 580 | AN | 3022 | THERE |
| 563 | NO | 2997 | *CAPTAIN* |
| 547 | WHEN | 2979 | WE |
| 511 | UPON | 2869 | SO |
| 502 | HAVE | 2635 | WHICH |
| 479 | ME | 2618 | SEA |
| 478 | WHAT | 2592 | HER |

| Simple network | | CHVG | |
|---|---|---|---|
| **Weight** | **Word** | **Weight** | **Word** |
| 467 | MORE | 2591 | OUT |
| 458 | OUT | 2590 | **SPERM** |
| 451 | WE | 2575 | HAVE |
| 445 | UP | 2538 | OLD |
| 441 | INTO | 2482 | THOU |
| 433 | THESE | 2351 | THEM |
| 431 | OLD | 2317 | **WHALES** |
| 429 | AHAB | 2291 | ONE |
| 425 | THEM | 2259 | ITS |
| 425 | ITS | 2252 | MAN |
| 414 | YE | 2214 | WHAT |
| 397 | YET | 2187 | **STARBUCK** |
| 381 | HER | 2159 | LIKE |
| 380 | WHO | 2085 | **WHITE** |
| 369 | OVER | 2053 | INTO |
| 361 | STILL | 2010 | MORE |
| 360 | THOUGH | 1981 | NO |
| 360 | ONLY | 1944 | THEN |
| 353 | MAN | 1934 | SOME |
| 352 | HERE | 1903 | UP |
| 351 | WILL | 1891 | AN |
| 348 | SEA | 1872 | UPON |
| 343 | SUCH | 1846 | THESE |
| 343 | LONG | 1836 | SUCH |
| 339 | VERY | 1788 | WHEN |
| 338 | WOULD | 1694 | BEEN |
| 336 | ABOUT | 1665 | **PEQUOD** |
| 331 | THOSE | 1634 | ABOUT |
| 326 | BEEN | 1592 | THOUGH |
| 321 | OTHER | 1589 | SEEMED |
| 320 | YOUR | 1574 | YOUR |
| 318 | THOU | 1549 | OVER |
| 317 | IF | 1544 | **OUR** |
| 316 | DOWN | 1540 | THOSE |
| 310 | ANY | 1540 | **DECK** |
| 307 | AFTER | 1521 | **HAS** |
| 306 | MOST | 1496 | **HEAD** |
| 304 | SHIP | 1491 | **MEN** |
| 303 | TWO | 1459 | MOST |
| 301 | THAN | 1446 | WILL |
| 301 | CHAPTER | 1443 | WOULD |
| 300 | BEFORE | 1428 | DOWN |
| 295 | GREAT | 1419 | **DO** |
| 294 | AGAIN | 1415 | **US** |
| 283 | SEEMED | 1415 | HERE |
| 283 | BEING | 1399 | GREAT |
| 280 | HOW | 1385 | YET |
| 279 | WHILE | 1357 | **SAID** |
| 275 | CAPTAIN | 1342 | VERY |
| 268 | STUBB | 1335 | ANY |

* The words present in the first one hundred of CHVG nodes but absent from the first one hundred of simple network nodes are in bold. The most informationally significant words from the CHVG top 100, which are also present in simple network top 100, are in italics.